\documentclass[lettersize,journal]{IEEEtran}
\usepackage{amsmath,amsfonts}
\usepackage{algorithmic}
\usepackage{algorithm}
\usepackage{array}
\usepackage{textcomp}
\usepackage{stfloats}
\usepackage[hyphens]{url}
\usepackage{verbatim}
\usepackage{graphicx}
\usepackage{cite}
\usepackage{bm}
\usepackage[hidelinks]{hyperref}
\usepackage{caption}
\usepackage{subcaption}
\usepackage{amsthm}
\usepackage{mathtools}
\usepackage[switch]{lineno}
\usepackage{makecell}
\usepackage{multirow}
\usepackage{mathdots}
\theoremstyle{definition}
\newtheorem{definition}{Definition}

\newcommand{\revision}[1]{{\color{black} #1}}

\usepackage[some]{background}
\SetBgScale{1}
\SetBgContents{\parbox{10cm}{%
  \Huge Draft:  \today\\[18cm]\rotatebox{180}{\Huge Draft:  \today}}}
\SetBgColor{red}
\SetBgAngle{270}
\SetBgOpacity{0.2}

\begin{document}

\title{Synchronized Object Detection for Autonomous Sorting, Mapping, and Quantification of \revision{Materials in Circular Healthcare}}

\author{Federico Zocco, Daniel R. Lake, Se\'{a}n McLoone, and Shahin Rahimifard
\thanks{This work has been conducted as part of the research project ‘Circular Economy for Small Medical Devices (ReMed)’, which is funded by the Engineering and Physical Sciences Research Council (EPSRC) of the UKRI (contract no: EP/W002566/1). The project funders were not directly involved in the writing of this article. For the purpose of open access, the author(s) has applied a Creative Commons Attribution (CC BY) license to any Accepted Manuscript version arising.}
\thanks{(\emph{Corresponding author: Federico Zocco})}
\thanks{Federico Zocco and Shahin Rahimifard are with the Centre for Sustainable Manufacturing and Recycling Technologies (SMART), Wolfson School of Mechanical, Electrical and Manufacturing Engineering, Loughborough University, England, UK (e-mail: federico.zocco.fz@gmail.com, s.rahimifard@lboro.ac.uk).}
\thanks{Daniel R. Lake is with the Intelligent Automation Centre (IAC), Wolfson School of Mechanical, Electrical and Manufacturing Engineering, Loughborough University, England, UK (e-mail: d.r.lake@lboro.ac.uk).}
\thanks{Se\'{a}n McLoone is with the Centre for Intelligent Autonomous Manufacturing Systems, School of Electronics, Electrical Engineering and Computer Science, Queen’s University Belfast, Northern Ireland, UK (e-mail: s.mcloone@qub.ac.uk).}
}

\maketitle

\begin{abstract} %
The circular economy paradigm is gaining interest as a solution to \revision{reducing} both material supply uncertainties and waste generation. One of the main challenges \revision{in realizing this paradigm} is monitoring materials, since in general, something that is not measured cannot be effectively managed. In this paper, we propose \revision{a real-time synchronized object detection framework that enables}, at the same time, autonomous sorting, mapping, and quantification of \revision{solid materials. We begin by introducing the general framework for real-time wide-area material monitoring, and then, we illustrate it using a numerical example. Finally, we develop a first prototype whose working principle is underpinned by the proposed framework. The prototype detects 4 materials from 5 different models of inhalers and, through a synchronization mechanism, it combines the detection outputs of 2 vision units running at 12-22 frames per second (Fig. \ref{fig:ForTitlePage}). This led us to introduce the notion of synchromaterial and to conceive a robotic waste sorter as a node compartment of a material network.} Dataset, code, and demo videos are publicly available\footnote{\url{https://github.com/fedezocco/2MMUsMed}}.  
\end{abstract}


\newcounter{mytempeqncnt}

\section{Introduction}
The UK alone generated 222.2 million tonnes of total waste in 2018 \cite{UKgov}, while each EU inhabitant generated 4.8 tonnes of waste in 2020 \cite{Eurostat}. The recycling rate is around 45\% in the UK \cite{UKgov} and 39.2\% in the EU \cite{Eurostat}. In addition to waste generation, several countries are experiencing uncertainties in material supplies, e.g., the UK \cite{CRM-UK} and the EU \cite{CRM-EU}. Some of these are critical materials that enable green and digital technologies such as lithium for electric vehicles, tungsten for mobile phones \cite{NHM}, and boron for wind turbines \cite{JRCboron}. Transitioning from a linear to a circular economy has the potential to reduce both waste generation and supply uncertainties by recovering and reusing materials as much as possible \cite{EllenMacArthur}.  

\revision{The increasing concern of the EU \cite{CRM-EU} and the UK \cite{CRM-UK} regarding the supply of critical raw materials (CRMs) is intensifying the need for wider adoption of the practices at the foundations of a circular economy \cite{EllenMacArthur}. Examples are reduce, reuse, rapair, recycle, and remanufacture practices \cite{morseletto2020targets} and their adoption is considered a measure to reduce the risk of CRM supply disruptions \cite{CRM-EU}. At the same time, circularity practices can reduce the rate of generation of waste since they can significantly extend the life of materials and products \cite{bakker2021understanding}.   
     
One of the main challenges for implementing a circular economy is monitoring materials and products, since in general, \emph{something that is not measured cannot be effectively managed}. This is true for the management of anything: of money, of water, and of the spread of a virus during a pandemic. To speed-up the transition to circularity, some approaches have been proposed for improving the monitoring of materials by leveraging computer vision. However, these mainly focus on the construction sector, do not work in real-time, and do not combine the measurements provided by multiple vision systems located at different places \cite{raghu2023towards,okuyamadeep,arbabi2022scalable,harrison2024scalability}.  
\begin{figure}%
\includegraphics[width=0.43\textwidth]{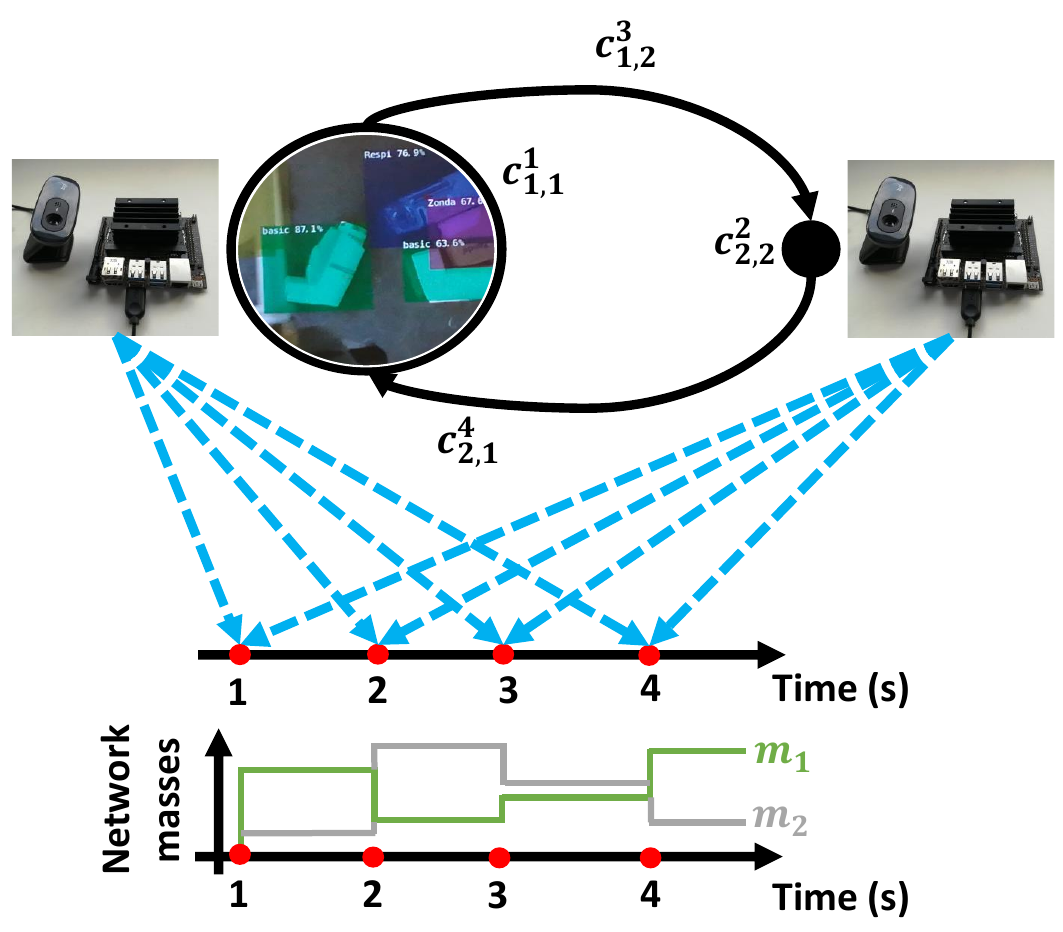}
\centering
\caption{High-level summary of the paper: real-time inhaler detection enables \revision{both material sorting, mapping, and quantification} when it is synchronized with other detection units.}
\label{fig:ForTitlePage}
\end{figure} 
  
In this context, the paper makes the following contributions: 
\begin{itemize}
\item{The general framework for real-time wide-area material mapping and quantification is given in Section \ref{sec:theory} and illustrated with a numerical example (Section \ref{sub:example1}).}
\item{A prototype system is developed and demonstrated for a medical device reprocessing application (Section \ref{sub:prototype}); its working principle is summarized in Fig. \ref{fig:ForTitlePage}.}
\item{The integration of the vision system for robotic waste sorting with the wide-area material monitoring system is highlighted as a means of achieving inter-operation between local and global material monitoring (Section \ref{sub:prototype}).}
\end{itemize}
Throughout the paper, sets are indicated with calligraphic upper case letters, while matrices and vectors are indicated with bold upper case and bold lower case letters, respectively.

The remainder of the paper is structured as follows. Section \ref{sec:RelatedWork} discusses the state-of-the-art of related work, Section \ref{sec:theory} provides the general framework for real-time wide-area material mapping and quantification, Section \ref{sec:examplesPrototype} covers a numerical example and a prototype, and finally Section \ref{sec:Conclusions} gives the conclusions.

\section{Related Work}\label{sec:RelatedWork}
\subsection{Towards Autonomous Material Monitoring for Circular Economy}
The transition to a circular economy requires materials and products to be monitored through their life-cycle more accurately and in a more timely fashion than it has been done to date \cite{fu2022evolution}. There are few existing works on the automation of material detection for the mapping and quantification of resources and, since these works target the construction sector, their focus is on building materials \cite{raghu2023towards,okuyamadeep,arbabi2022scalable,harrison2024scalability}. In contrast, our system targets materials that can be found in products located inside buildings, and hence, are not amenable to satellite-image-based tracking approaches as explored by Ragu \emph{et al.} \cite{raghu2023towards} and Okuyama \cite{okuyamadeep}.  

At a methodological level, our approach provides the spatial localization of materials similar to Ragu \emph{et al.} \cite{raghu2023towards} and Arbabi \emph{et al.} \cite{arbabi2022scalable}, and estimates the quantity of each type of material as achieved by Okuyama \cite{okuyamadeep} using Street View images. In addition, we also perform a granular per-image detection similar to Arbabi \emph{et al.} \cite{arbabi2022scalable}, which uses image segmentation rather than classification as employed in \cite{raghu2023towards,okuyamadeep,harrison2024scalability}. However, differently from Arbabi \emph{et al.} \cite{arbabi2022scalable}, we opted for object detection (covered in the next section) rather than image segmentation in order to reduce the computational load, and hence, latency. Reducing latency is important in our system because, in contrast to other works \cite{raghu2023towards,okuyamadeep,arbabi2022scalable,harrison2024scalability}, our prototype runs in real-time, i.e., it performs the detections with a speed of 12-22 frames per second and, every 1 second, it updates the material stock estimates. 

Harrison \emph{et al.} \cite{harrison2024scalability} generated and tested a synthetic dataset for training classifiers since the manual annotation of images is time consuming; in this work, we use manual annotation. Automating the annotation process for object detection is more complex than automating it for image classification.   

Another difference from existing works \cite{raghu2023towards,okuyamadeep,arbabi2022scalable,harrison2024scalability} is that our system leverages a synchronization mechanism that combines the local detections coming from multiple vision units to give global estimates of stocks. Finally, while Okuyama \cite{okuyamadeep} gives the equations for estimating the building materials from images, we give the equations for any class of material considering multiple synchronized vision units, which can serve as a general framework for real-time wide-area autonomous material mapping and quantification.

\subsection{Object Detection}
Object detection is one of the fundamental tasks in computer vision. Essentially, it consists of extracting information about the classes and locations of the objects depicted in images or video frames. The history of object detection is typically divided into two major periods. The first period ended approximately in 2014 and it comprised algorithms based on manual designs of features. The second period is on-going and is based on the end-to-end deep learning of features \cite{zou2023object}.

Two of the most popular object detectors are the you-only-look-once (YOLO) detector \cite{redmon2016you} and the single-shot multibox detector (SSD) \cite{liu2016ssd} because of their good speed-accuracy trade-off. Another class of detectors are the EfficientDet detectors \cite{tan2020efficientdet}, whose efficiency was further improved by Zocco \emph{et al.} \cite{zocco2023towards} by increasing the number of convolutional layers in the class/box subnets and reducing the number of bi-directional feature pyramid network layers. Pu \emph{et al.} \cite{pu2023adaptive} proposed an adaptive rotated convolution to improve the detection of objects with arbitrary orientation and they integrated the new convolution into the vision backbone networks.    

In the last four or five years, important results have been achieved with neural models based on transformers \cite{zou2023object}. A significant step in this direction was made in 2020 by Carion \emph{et al.} \cite{carion2020end}, where the Hungarian loss was used to evaluate the class and box predictions; the proposed architecture was a sequence of a convolutional network backbone, a transformer encoder, a transformer decoder, and feed forward networks as the prediction heads. Recently, the real-time detection transformer of Zhao \emph{et al.} \cite{zhao2024detrs} outperformed several versions of YOLO. Other recent works on transformer-based detection are \cite{li2022dn,li2023lite,liudab}.

The advancements in computing platforms optimized for training and inference with deep neural networks has led to the integration of graphics processing units (GPUs) and tensor processing units (TPUs) into light and affordable microcomputers suitable for the design of intelligent robotic and autonomous systems. In particular, for this paper we use two NVIDIA Jetson Nano developer kits, which are small GPU-based platforms announced in 2019 \cite{NvidiaNano2019}. Two more powerful (and more expensive) devices in the NVIDIA Jetson family are the Jetson Xavier and the Jetson Orin developer kits \cite{e-conSystemPage}. Running object detection on these platforms is possible with the library developed by NVIDIA \cite{NvidiaODtutorial}. Another edge device optimized for AI is the Coral Dev board developed by Google Research for adapting to portable microcomputers machine learning algorithms executed on hyperscalers \cite{CoralWebsite}; the Coral Dev board is based on the TPU technology instead of the GPU. Object detection can be run on the Coral Dev board as described in the Coral documentation \cite{CoralDevBoard}. A comparison of three edge devices for object detection was carried out by Kang \emph{et al.} \cite{kang2022evaluation}; they found that the NVIDIA Jetson Xavier NX performed the best overall, but it is four times more costly than the other platforms considered; moreover, the Jetson Nano and the Coral Dev Board Mini kits showed a similar latency, while the former is better than the latter in terms of accuracy.

\section{Wide-Area Material Monitoring}\label{sec:theory}
In this section, we first formally define what we mean by a material network. Then, we show how to monitor such networks. The analogy with electrical networks is essential to understand our idea, and hence, we will recall the analogy between material and electrical networks throughout the paper.
\subsection{Thermodynamical Material Networks}
As discussed in the introduction section, the design of circular flows of materials is an evolving area of research. To aid in this endeavor, recently, the formalism of thermodynamical material networks (TMNs) was proposed by Zocco \emph{et al.} \cite{zocco2023thermodynamical,zocco2022circularity,zocco2024unification} by generalizing the design approach of thermodynamic cycles, which is possible thanks to the generality of thermodynamics \cite{haddad2017thermodynamics}. Here, we briefly recall their definition and direct the interested reader to \cite{zocco2023thermodynamical,zocco2022circularity,zocco2024unification} for further details. 
\begin{definition}[\hspace{1sp}\cite{bondy1976graph}]\label{def:Digraph}
A directed graph $D$ or \emph{digraph} is a graph identified by a set of $n_\text{v}$ \emph{nodes} $\{v_1, v_2, \dots, v_{n_\text{v}}\}$ and a set of $n_\text{a}$ \emph{arcs} $\{a_1, a_2, \dots, a_{n_\text{a}}\}$ that connect the nodes. A digraph $D$ in which each node or arc is associated with a \emph{weight} is a \emph{weighted digraph}. 
\end{definition}
\begin{definition}[\hspace{1sp}\cite{zocco2023thermodynamical}]\label{def:TMN}
A \emph{thermodynamical material network} (TMN) is a set $\mathcal{N}$ of connected thermodynamic compartments, that is, 
\begin{equation}\label{def:TMNset}
\begin{gathered}
\mathcal{N} = \left\{c^1_{1,1}, \dots, c^{k_\text{v}}_{k_\text{v},k_\text{v}}, \dots, c^{n_\text{v}}_{n_\text{v},n_\text{v}}, \right. \\ 
\left. c^{n_\text{v}+1}_{i_{n_\text{v}+1},j_{n_\text{v}+1}}, \dots, c^{n_\text{v}+k_\text{a}}_{i_{n_\text{v}+k_\text{a}},j_{n_\text{v}+k_\text{a}}}, \dots, c^{n_\text{c}}_{i_{n_\text{c}},j_{n_\text{c}}}\right\}, 
\end{gathered}
\end{equation}
which transport, store, use, and transform a target material. Each compartment is indicated by a \emph{control surface} \cite{moran2010fundamentals} and is modeled using \emph{dynamical systems} derived from a mass balance and/or at least one of the laws of \emph{thermodynamics} \cite{haddad2019dynamical}.
\end{definition}

Specifically, $\mathcal{N} = \mathcal{R} \cup \mathcal{T}$, where $\mathcal{R} \subseteq \mathcal{N}$ is the subset of compartments $c^k_{i,j}$ that \emph{store}, \emph{transform}, or \emph{use} the target material, while $\mathcal{T} \subset \mathcal{N}$ is the subset of compartments $c^k_{i,j}$ that \emph{move} the target material between the compartments belonging to $\mathcal{R} \subseteq \mathcal{N}$. A net $\mathcal{N}$ is associated with its weighted \emph{mass-flow digraph} $M(\mathcal{N})$, which is a weighted digraph whose nodes are the compartments $c^k_{i,j} \in \mathcal{R}$ and whose arcs are the compartments $c^k_{i,j} \in \mathcal{T}$. For node-compartments $c^k_{i,j} \in \mathcal{R}$ it holds that $i = j = k$, whereas for arc-compartments $c^k_{i,j} \in \mathcal{T}$ it holds that $i \neq j$ because an arc moves the material from the node-compartment $c^i_{i,i}$ to the node-compartment $c^j_{j,j}$. The orientation of an arc is given by the direction of the material flow. The superscript $k$ is the identifier of each compartment. The weight assigned to a node-compartment is the mass stock $m_k$ within the corresponding compartment, whereas the weight assigned to an arc-compartment is the mass flow rate $\dot{m}_{i,j}$ from the node-compartment $c^i_{i,i}$ to the node-compartment $c^j_{j,j}$. The superscripts $k_\text{v}$ and $k_\text{a}$ in (\ref{def:TMNset}) are the $k$-th node and the $k$-th arc, respectively, while $n_\text{c}$ and $n_\text{v}$ are the total number of compartments and nodes, respectively. Since $n_\text{a}$ is the total number of arcs, it holds that $n_\text{c} = n_\text{v} + n_\text{a}$ \cite{zocco2023thermodynamical,zocco2022circularity,zocco2024unification}.

The digraph of a TMN is shown on the left-hand side of Fig. \ref{fig:ElectricalAndMaterialNets}, where the node and the arc compartments are indicated. In this case, 
\begin{equation}
\begin{aligned}
\mathcal{N} = \{c^1_{1,1}, c^2_{2,2}, c^3_{3,3}, c^4_{4,4}, c^5_{5,5}, c^6_{6,6}, c^7_{7,7}, c^8_{8,8}, \\ c^9_{9,9}, c^{10}_{1,2}, c^{11}_{2,3}, c^{12}_{3,4}, c^{13}_{3,5}, c^{14}_{5,6}, c^{15}_{6,7}, c^{16}_{7,8}, c^{17}_{9,6}\},
\end{aligned}
\end{equation}
$n_\text{v} = 9$, $n_\text{a} = 8$, and $n_\text{c} = 17$. On the right-hand side of the same figure (i.e., Fig. \ref{fig:ElectricalAndMaterialNets}), an electrical network is depicted to show that the two systems are similar when modeled as a digraph \cite{balabanian1969electrical}. While solid materials flow in the former (e.g., via transportation by truck), charged particles flow in the latter. Both systems have source nodes (material reservoirs are in the former). The retailers and users of a material network can be seen as the distribution and the load nodes of an electrical network, respectively.        
\begin{figure}
\includegraphics[width=0.45\textwidth]{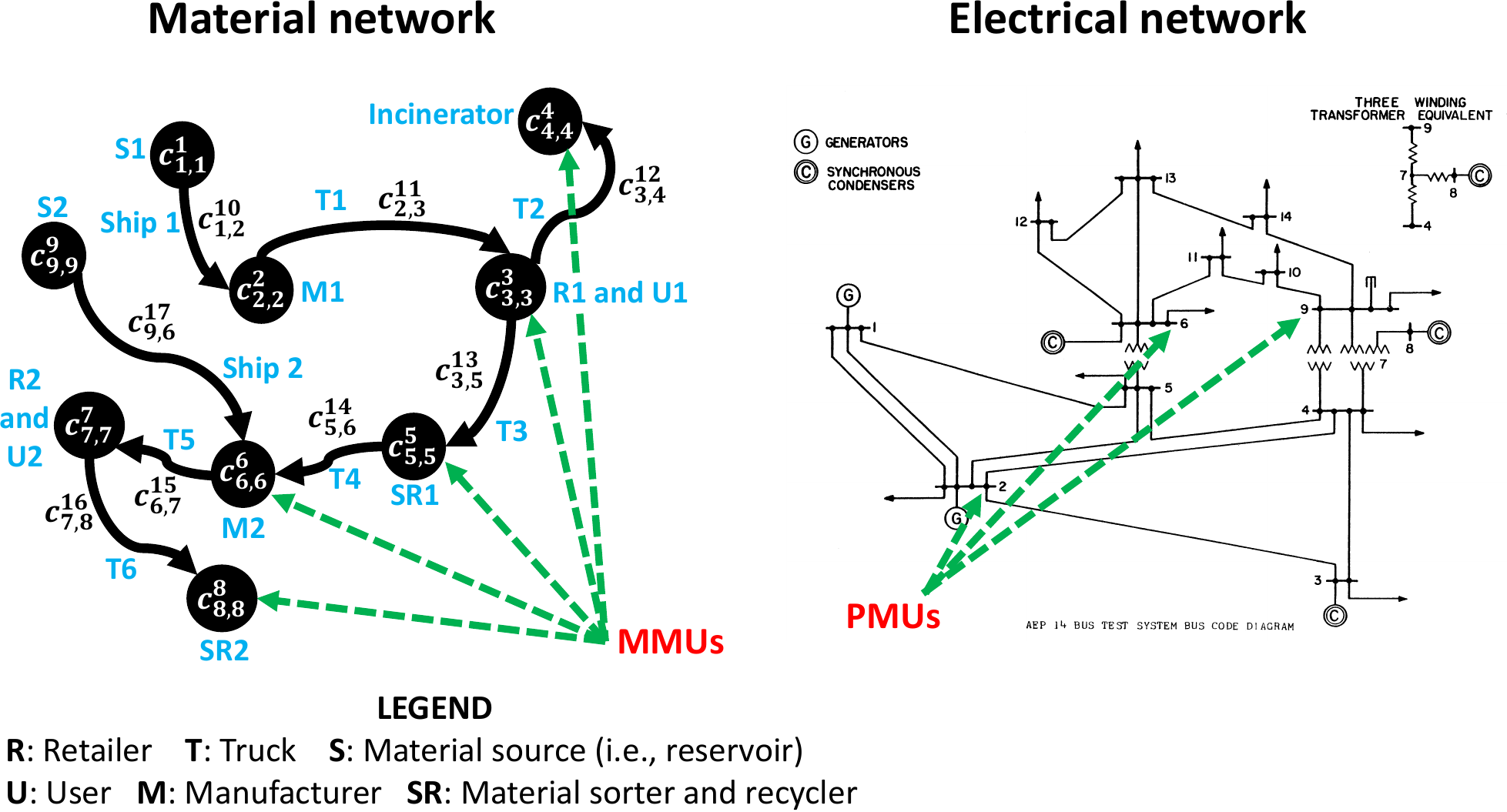}
\centering
\caption{\revision{Analogy between a material network (left) and an electrical network (right). Wide-area monitoring of these networks is performed via material measurement units (MMUs) \cite{zocco2022material} and PMUs \cite{khajeh2015integrated}, respectively.}}
\label{fig:ElectricalAndMaterialNets}
\end{figure} 

The next section will answer the question ``How to monitor in real-time the state of $\mathcal{N}$, i.e., the material mass $m_k$ within the $k$-th node?''. Since the transition to a more circular economy is accelerated by improving the mapping and quantification of materials and products over their life-cycle \cite{fu2022evolution}, we observed how a modern electrical network monitors its highly dynamic state. It does so using phasor measurement units (PMUs) and synchrophasors \cite{khajeh2015integrated}. Hence, to monitor in real-time the material mass $m_k$ inside the $k$-th thermodynamic compartment, as asked in the question above, we could adapt the working principle of PMUs to a material network $\mathcal{N}$. This led to the idea of material measurement units (MMUs) discussed in \cite{zocco2022material} without examples or implementations, which are provided in this paper (Section \ref{sec:examplesPrototype}). The next section will give the underpinning framework.

\subsection{Framework for Wide-Area Material Monitoring}
The material stock $m_k$ within the $k$-th node can be automatically estimated in real-time using multiple vision systems as follows. Let $\mathcal{U}$ be the set of identifiers $k$ of the compartments $c^k_{i,j}$ where a unit (i.e., an MMU) is placed; thus, $|\mathcal{U}| = S$ is the number of units. The left-hand side of Fig. \ref{fig:ElectricalAndMaterialNets} is an example with $\mathcal{U} = \{3, 4, 5, 6, 8\}$ and $S = 5$. Let $\mathcal{C}_k(n) \subset \mathbb{N}$ be the set of object classes \emph{detected} by the $k$-th unit at time $n$, with $|\mathcal{C}_k(n)| = Q_k(n)$ and with $q \in \mathcal{C}_k(n)$ a detected class; let $\mathcal{M}_k(n) \subset \mathbb{N}$ be the set of constituent materials of the object classes $\mathcal{C}_k(n)$ detected by the $k$-th unit at time $n$, with $|\mathcal{M}_k(n)| = \Psi_k (n)$, with $\psi \in \mathcal{M}_k(n; q)$ a constituent material of the detected object class $q$, and with 
\begin{equation}
\mathcal{M}_k(n) = \bigcup\limits_{q} \mathcal{M}_k(n; q).
\end{equation}
Hence, the measurement $\hat{m}_k (n)$ of the mass $m_k$ at time $n$ is
\begin{equation}\label{eq:massFromMMU}
\hat{m}_k (n) = \sum\limits_{q \, \in \, \mathcal{C}_k(n)} \,\, \sum\limits_{\psi \, \in \, \mathcal{M}_k(n; q)}  f^\psi_{k,q} = \sum\limits_{\psi \, \in \, \mathcal{M}_k(n)}  \hat{F}^\psi_{k}(n),
\end{equation}    
where $n \in \overline{\mathbb{Z}}_+$ is the index of the sample, $\overline{\mathbb{Z}}_+$ is the set of nonnegative integers, $f^\psi_{k,q}$ is the fraction of mass of material $\psi$ contained in the object class $q$ detected by the unit $k$ inside the compartment $k$ at time $n$, and where $\hat{F}^\psi_{k}(n)$ is the mass of the material $\psi$ detected by the unit $k$ inside the compartment $k$ at time $n$, i.e.,
\begin{equation}\label{eq:Fkpsi}
\hat{F}^\psi_{k}(n) = \sum\limits_{q \, \in \, \mathcal{C}_k(n)} f^\psi_{k,q}.
\end{equation}    
Thus, the total mass measured by the set of units $\mathcal{U}$ at time $n$ is given by
\begin{equation}\label{eq:totalMass}
\hat{l}(n) = \sum\limits_{k \, \in \, \mathcal{U}} \hat{m}_k (n) = \sum\limits_{k \, \in \, \mathcal{U}} \,\, \sum\limits_{q \, \in \, \mathcal{C}_k(n)} \,\, \sum\limits_{\psi \, \in \, \mathcal{M}_k(n; q)}  f^\psi_{k,q}.
\end{equation}  
Another useful quantity to define is the total mass of material $\psi$ detected by the set of units $\mathcal{U}$, i.e.,
\begin{equation}\label{eq:Fpsi}
\hat{F}^\psi(n) = \sum\limits_{k \, \in \, \mathcal{U}} \hat{F}^\psi_{k}(n) = \bm{1}^\top \hat{\bm{F}}^\psi(n) \,\bm{1}.
\end{equation} 
Here, $\bm{1} \in \mathbb{R}^{n_\text{v}}$ is a vector of ones and $\hat{\bm{F}}^\psi(n) \in \mathbb{R}^{n_\text{v} \times n_\text{v}}$ is a matrix containing the information on the distribution of  $\psi$ across the compartments as follows: its diagonal entries are $\hat{F}^\psi_{k}(n) \ge 0$ for $k \in \mathcal{U}$ and for $k$ such that $c^k_{i,j} \in \mathcal{R}$ (i.e., nodes with MMUs); its off-diagonal entries are $\hat{F}^\psi_{k}(n) \ge 0$ for $k \in \mathcal{U}$, for $k$ such that $c^k_{i,j} \in \mathcal{T}$ (i.e., arcs with MMUs), and they are located in row $i$ and column $j$; finally, for all the entries with $k \notin \mathcal{U}$, it holds that $\hat{F}^\psi_{k}(n) = 0$ (i.e., where no MMU is placed). 

Furthermore, by removing the condition $k \in \mathcal{U}$ from the definition of $\hat{\bm{F}}^\psi(n)$ (which selects only the compartments with MMUs), we obtain the matrix $\bm{F}^\psi(n) \in \mathbb{R}^{n_\text{v} \times n_\text{v}}$, whose entries are the actual masses of material $F^\psi_{k}(n)$ in each compartment in the network, including those that are not measured by MMUs; $F^\psi_{k}(n) = 0$ where there is no arc.     

For example, in the case of the material network on the left-hand side of Fig. \ref{fig:ElectricalAndMaterialNets}, $\bm{F}^\psi(n)$ and $\hat{\bm{F}}^\psi(n)$ take the form given in equations (\ref{eq:matrixF-example}) and (\ref{eq:matrixFhat-example}), respectively. In general, the sparsity of $\hat{\bm{F}}^\psi(n)$ decreases by increasing the number of MMUs.
\begin{figure*}[!t]
\normalsize
\setcounter{mytempeqncnt}{\value{equation}}
\revision{
\begin{equation}
\label{eq:matrixF-example} 
\bm{F}^\psi(n)=
\begin{bmatrix}
F^\psi_{1}(n) & F^\psi_{10}(n) & 0 & 0 & 0 & 0 & 0 & 0 & 0\\
0 & F^\psi_{2}(n) & F^\psi_{11}(n) & 0 & 0 & 0 & 0 & 0 & 0\\
0 & 0 & F^\psi_{3}(n) & F^\psi_{12}(n) & F^\psi_{13}(n) & 0 & 0 & 0 & 0\\ 
0 & 0 & 0 & F^\psi_{4}(n) & 0 & 0 & 0 & 0 & 0\\ 
0 & 0 & 0 & 0 & F^\psi_{5}(n) & F^\psi_{14}(n) & 0 & 0 & 0\\ 
0 & 0 & 0 & 0 & 0 & F^\psi_{6}(n) & F^\psi_{15}(n) & 0 & 0\\ 
0 & 0 & 0 & 0 & 0 & 0 & F^\psi_{7}(n) & F^\psi_{16}(n) & 0 \\ 
0 & 0 & 0 & 0 & 0 & 0 & 0 & F^\psi_{8}(n) & 0\\ 
0 & 0 & 0 & 0 & 0 & F^\psi_{17}(n) & 0 & 0 & F^\psi_{9}(n)\\ 
\end{bmatrix}
\end{equation}
\begin{equation}
\label{eq:matrixFhat-example} 
\hat{\bm{F}}^\psi(n)=
\begin{bmatrix}
0 & 0 & 0 & 0 & 0 & 0 & 0 & 0 & 0\\
0 & 0 & 0 & 0 & 0 & 0 & 0 & 0 & 0\\
0 & 0 & \hat{F}^\psi_{3}(n) & 0 & 0 & 0 & 0 & 0 & 0\\ 
0 & 0 & 0 & \hat{F}^\psi_{4}(n) & 0 & 0 & 0 & 0 & 0\\ 
0 & 0 & 0 & 0 & \hat{F}^\psi_{5}(n) & 0 & 0 & 0 & 0\\ 
0 & 0 & 0 & 0 & 0 & \hat{F}^\psi_{6}(n) & 0 & 0 & 0\\ 
0 & 0 & 0 & 0 & 0 & 0 & 0 & 0 & 0 \\ 
0 & 0 & 0 & 0 & 0 & 0 & 0 & \hat{F}^\psi_{8}(n) & 0\\ 
0 & 0 & 0 & 0 & 0 & 0 & 0 & 0 & 0\\ 
\end{bmatrix}
\end{equation}
}
\hrulefill
\vspace*{4pt}
\end{figure*}

Now, note that the $S$ MMU measurements $\hat{m}_k (n)$ must be provided at every time step $n$ to compute (\ref{eq:totalMass}), where $\hat{m}_k (n)$ and $F_k^\psi(n)$ are the measurements taken by the $k$-th unit (see (\ref{eq:massFromMMU})). This means that the $S$ measurements provided by the $S$ MMUs must be \emph{synchronized in time} to compute equations (\ref{eq:totalMass}) and (\ref{eq:Fpsi}). To highlight the synchronization requirement, we introduce the notion of \emph{synchromaterial} (note the analogy with the \emph{synchrophasor} measured by PMUs \cite{dwivedi2023dynamopmu}).
\begin{definition}[Synchromaterial]\label{def:synchromat}
Mass of a material automatically detected at one or more locations and computed in a synchronized fashion using material attributes.
\end{definition}
Examples of the material attributes in Definition \ref{def:synchromat} are the material mass, the material type, and the class of the product containing the material of interest. Indeed, in this paper, a product class is conceived as an attribute of the constituent materials because the product class (e.g., a bottle, a laptop) defines the specific functionality and shape of the constituent materials (e.g., the functionality and shape of the constituent materials of a bottle and a laptop are indicated by the product class ``bottle'' and ``laptop'', respectively). 

From Definition \ref{def:synchromat} it follows that the total detected masses (\ref{eq:totalMass}) and (\ref{eq:Fpsi}) are synchromaterials computed through the synchronized measurements provided by the set of units $\mathcal{U}$. In contrast, the masses given by the $k$-th unit (\ref{eq:massFromMMU})-(\ref{eq:Fkpsi}) may or may not be a synchromaterial and this depends on the chosen system setup. Indeed, in general, $\hat{m}_k(n)$ and $F^\psi_{k}(n)$ can be computed independently of the other units.   

So far, we introduced the notion of a TMN as a material network $\mathcal{N}$ (Eq. (\ref{def:TMNset})) and its analogy with electrical networks (Fig. \ref{fig:ElectricalAndMaterialNets}). Then, we showed how to monitor in real-time the state of $\mathcal{N}$, i.e., the material mass $m_k$ inside the $k$-th thermodynamic compartment, by leveraging the working principle of PMUs in modern electrical networks. Specifically, we formulated the calculation of $m_k$ given by a unit (i.e., an MMU) placed in the $k$-th compartment (see Equation (\ref{eq:massFromMMU})), and then, we introduced the notion of synchromaterial and the calculation of the total detected masses (\ref{eq:totalMass})-(\ref{eq:Fpsi}), which are synchromaterials. The name ``synchromaterial'' is motivated by the analogy with the synchronization mechanism of synchrophasors. 

The next section will cover a numerical example to illustrate the calculation of the synchromaterials using object detectors. The working principle is valid regardless of the detector in use (e.g., YOLO, SSD, EfficientDet). However, different detectors may provide different detection performance, i.e., different $\mathcal{C}_k (n)$, which directly affects the calculation of the compartment-level mass (\ref{eq:massFromMMU}), and then, it propagates to the calculation of the network masses (\ref{eq:totalMass})-(\ref{eq:Fpsi}). In particular, misdetections and missed detections increase the measurement error inside the $k$-th node 
\begin{equation}
e_k(n) = m_k(n) - \hat{m}_k(n),    
\end{equation}
where $m_k(n)$ is the mass inside the $k$-th node at time $n$.

\section{Example and Prototype}\label{sec:examplesPrototype}
This section illustrates the framework covered in the previous section using a numerical example (Section \ref{sub:example1}), and then, it describes a first prototype (Section \ref{sub:prototype}).

\subsection{Numerical Example: Two Units, Two Object Classes, Two Materials, and Perfect Detections}\label{sub:example1}
Consider the material network in Fig. \ref{fig:consideredTMN}, with $\mathcal{N} = \{c^1_{1,1}, c^2_{2,2}, c^3_{1,2}, c^4_{2,1}\}$, $n_\text{v} = 2$, $n_\text{a} = 2$, and $n_\text{c} = 4$. The arcs are the trucks moving a mass $\overline{m}$ between the nodes.
\begin{figure}
\includegraphics[width=0.45\textwidth]{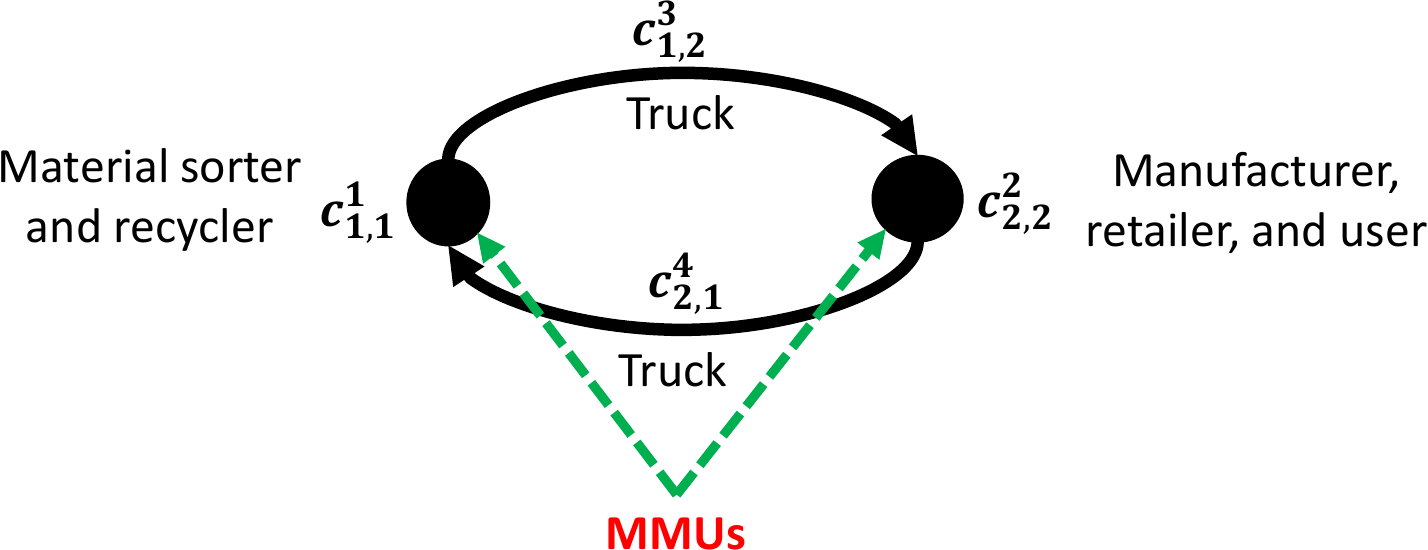}
\centering
\caption{\revision{Material network considered in the example and the prototype.}}
\label{fig:consideredTMN}
\end{figure}
The dynamics of the masses (i.e., stocks) inside the nodes are \cite{zocco2022circularity} 
\begin{equation}
\begin{gathered}\label{eq:mass1}
m_1(n+1) = m_1(n) + \overline{m} \left(\delta_{t_{1,\text{in},4}}(t) \right. \\ \left. - \delta_{t_{1,\text{out},3}}(t) \right), \quad n \in \overline{\mathbb{Z}}_+,
\end{gathered}
\end{equation}
\begin{equation}\label{eq:mass2}
\begin{gathered}
m_2(n+1) = m_2(n) + \overline{m} \left(\delta_{t_{2,\text{in},3}}(t) \right. \\ \left. - \delta_{t_{2,\text{out},4}}(t) \right),\quad n \in \overline{\mathbb{Z}}_+,
\end{gathered}
\end{equation}
where it holds that $t = nT$, with $T$ the sample time. In practice, $T$ is chosen significantly smaller than the time needed for transportation and storage so that their dynamics are captured by this model, e.g., $T \leq 1$ minute. In addition, $\delta_{t_*}(t)$ is a rectangular pulse of short duration $\varepsilon$ emulating a unit impulse centered in $t_*$ and defined as
\begin{equation}
\delta_{t_*}(t) = \text{rect}\left(\frac{t - t_*}{\varepsilon}\right),
\end{equation}
where $\text{rect}(\cdot)$ is the rectangular function defined as \cite{vitettaTdS}
\begin{equation}
\text{rect}(\sigma) = 
\begin{dcases*}
1 & if $|\sigma| < 1/2$ \\
1/2 & if $|\sigma| = 1/2$ \\
0 & otherwise. \\
\end{dcases*}
\end{equation}
Furthermore, $t_{i,\text{in},j}$ is the time instant at which the mass $\overline{m}$ enters the compartment $i$ from the compartment $j$, while $t_{i,\text{out},j}$ is the time instant at which the mass $\overline{m}$ exits the compartment $i$ for the compartment $j$. Note that, for the physical consistency of the equations, the following equalities between the time instants hold:
\begin{equation}\label{eq:cs-constraints}
t_{i,\text{in},j} = t_{j,\text{out},i}, \quad i, j \in \{1, 2, 3, 4\}, \, i \neq j.
\end{equation} 
Finally, without loss of generality, we consider the case of $t_{1,\text{in},4} > t_{2,\text{out},4} > t_{2,\text{in},3} > t_{1,\text{out},3}$. 

As indicated in Fig. \ref{fig:consideredTMN}, two units are used for monitoring the dynamics of the masses: one unit in $c^1_{1,1}$ and one unit in $c^2_{2,2}$; thus, $\mathcal{U} = \{1, 2\}$ and $S = 2$. Consider that $\mathcal{N}$ contains two types of objects: the first type has a mass of 20 grams and it is made of 75\% rubber and 25\% plastic; the second type has a mass of 80 grams and is made of 50\% rubber and 50\% plastic. Consider also that an object of the second type is being transported from $c^1_{1,1}$ to $c^2_{2,2}$ and then back from $c^2_{2,2}$ to $c^1_{1,1}$; thus, $\overline{m} = 80$ grams (see (\ref{eq:mass1})-(\ref{eq:mass2})). The transported mass $\overline{m}$ enters and exits the nodes at the following times: $t_{1,\text{out},3} = 10$ h, $t_{2,\text{in},3} = 30$ h, $t_{2,\text{out},4} = 40$ h, and $t_{1,\text{in},4} = 60$ h. Finally, consider that $c^1_{1,1}$ initially contains an object of the first type and one object of the second type, while $c^2_{2,2}$ initially contains only one object of the second type. With this situation, let us see how the set of units $\mathcal{U}$ monitors the dynamics of the material masses in $\mathcal{N}$ in the case of perfect detections, i.e., without misdetections and without missed detections (which can occur with real object detection algorithms).

First of all, let us assign a class index to each type of object, which would be the class of the object used by the object detection algorithm if the system is implemented. The choice of the classes is arbitrary, but each class must be unique for each type of object. Hence, we assign the class index 2 to the first type of object and the class index 3 to the second type of object. This yields $\mathcal{C}_1(0) = \{2, 3\}$ and $\mathcal{C}_2(0) = \{3\}$. Let us now assign a unique index to each type of constituent material of the objects. In this case, the constituent materials of the object types are plastic and rubber; thus, we arbitrarily choose the index 4 for the rubber and the index 7 for the plastic. This yields $\mathcal{M}_1(0) = \mathcal{M}_2(0) = \{4, 7\}$. In addition, from the information about the material composition of the objects, we have that $f^4_{1,2} = 15$ grams, $f^7_{1,2} = 5$ grams, $f^4_{1,3} = 40$ grams, $f^7_{1,3} = 40$ grams,  $f^4_{2,2} = 0$ grams, $f^7_{2,2} = 0$ grams, $f^4_{2,3} = 40$ grams, and $f^7_{2,3} = 40$ grams. Note that $f^4_{2,2}$ and $f^7_{2,2}$ equal zero because the unit with $k = 2$ never detects the object class 2; only objects of class 3 appear in $c^2_{2,2}$.

In the case of perfect detections, the measurements provided by the set of units $\mathcal{U}$ are shown in Fig. \ref{fig:Example1} and given in the following equations.    
\begin{figure}
\includegraphics[width=0.45\textwidth]{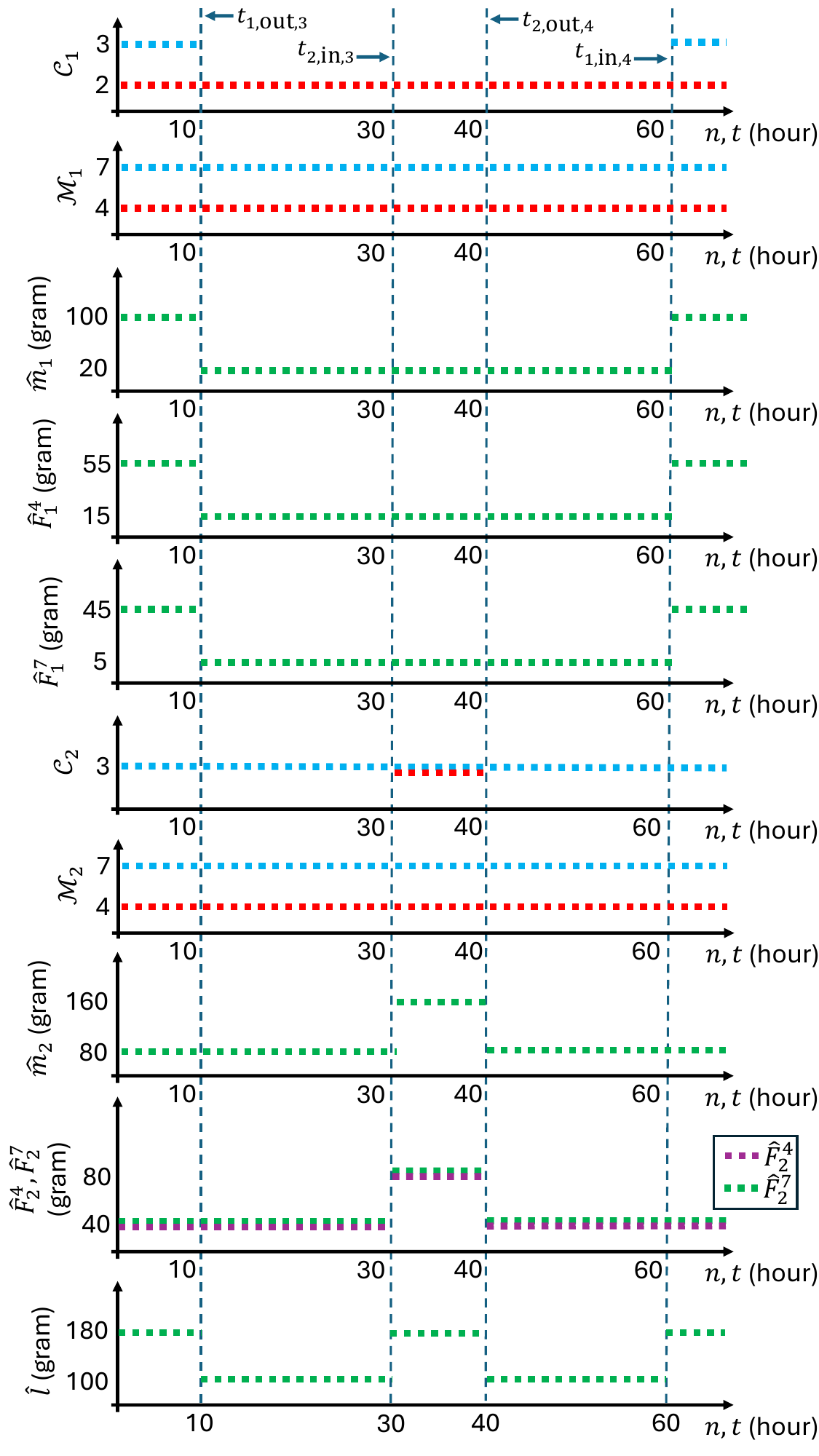}
\centering
\caption{\revision{Synchromaterials of the numerical example.}}
\label{fig:Example1}
\end{figure} 
\begin{equation}\label{eq:C1}
\mathcal{C}_1(n) = 
\begin{dcases*}
\{2, 3\} & if $0 \leq t \leq t_{1,\text{out},3}$\\
& or $t > t_{1,\text{in},4}$ \\
\{2\} & otherwise,
\end{dcases*}
\end{equation}
\begin{equation}
\mathcal{M}_1(n) = \{4, 7\} \quad \forall t,
\end{equation}
\begin{equation}
\hat{m}_1(n) =
\begin{dcases*}
f^4_{1,2} + f^7_{1,2} \\ + f^4_{1,3} + f^7_{1,3} & if  $0 \leq t \leq t_{1,\text{out},3}$\\
& or $t > t_{1,\text{in},4}$\\
f^4_{1,2} + f^7_{1,2} & otherwise,
\end{dcases*}
\end{equation}
\begin{equation}
\hat{F}^4_1(n) =
\begin{dcases*}
f^4_{1,2} + f^4_{1,3} & if $0 \leq t \leq t_{1,\text{out},3}$\\
& or $t > t_{1,\text{in},4}$\\
f^4_{1,2} & otherwise,
\end{dcases*}
\end{equation}
\begin{equation}
\hat{F}^7_1(n) =
\begin{dcases*}
f^7_{1,2} + f^7_{1,3} & if $0 \leq t \leq t_{1,\text{out},3}$\\
& or $t > t_{1,\text{in},4}$\\
f^7_{1,2} & otherwise,
\end{dcases*}
\end{equation}
\begin{equation}
\mathcal{C}_2(n) = 
\begin{dcases*}
\{3\} & if $0 \leq t \leq t_{2,\text{in},3}$\\
& or $t > t_{2,\text{out},4}$\\
\{3, 3\} & otherwise,
\end{dcases*}
\end{equation}
\begin{equation}
\mathcal{M}_2(n) = \{4, 7\} \quad \forall t,
\end{equation}
\begin{equation}
\hat{m}_2(n) =
\begin{dcases*}
f^4_{2,3} + f^7_{2,3} & if  $0 \leq t \leq t_{2,\text{in},3}$\\
& or $t > t_{2,\text{out},4}$\\
2\left(f^4_{2,3} + f^7_{2,3}\right) & otherwise,
\end{dcases*}
\end{equation}
\begin{equation}
\hat{F}^4_2(n) =
\begin{dcases*}
f^4_{2,3} & if $0 \leq t \leq t_{2,\text{in},3}$\\
& or $t > t_{2,\text{out},4}$\\
2f^4_{2,3} & otherwise,
\end{dcases*}
\end{equation}
\begin{equation}
\hat{F}^7_2(n) =
\begin{dcases*}
f^7_{2,3} & if $0 \leq t \leq t_{2,\text{in},3}$\\
& or $t > t_{2,\text{out},4}$\\
2f^7_{2,3} & otherwise,
\end{dcases*}
\end{equation}
\begin{equation}\label{eq:F-4}
\hat{F}^4(n) = \hat{F}^4_1(n) + \hat{F}^4_2(n),
\end{equation}
\begin{equation}\label{eq:F-7}
\hat{F}^7(n) = \hat{F}^7_1(n) + \hat{F}^7_2(n),
\end{equation}
\begin{equation}
\hat{\bm{F}}^4(n) =
\begin{bmatrix}
\hat{F}^4_1(n) & 0 \\
0 & \hat{F}^4_2(n) \\
\end{bmatrix},
\end{equation}
\begin{equation}
\hat{\bm{F}}^7(n) =
\begin{bmatrix}
\hat{F}^7_1(n) & 0 \\
0 & \hat{F}^7_2(n) \\
\end{bmatrix},
\end{equation}
and finally, 
\begin{equation}\label{eq:l-hat}
\hat{l}(n) = \hat{m}_1(n) + \hat{m}_2(n).
\end{equation}

Let us now analyze Fig. \ref{fig:Example1} along with the associated Equations (\ref{eq:C1})-(\ref{eq:l-hat}). All the variables are recorded at every time step $n$ (a physical implementation of the system would require a synchronization mechanism) and they show jump discontinuities when the mass $\overline{m}$ (i.e., an object of class 3) exits or enters a node, i.e., for $t = t_{1,\text{out},3}, t_{2,\text{in},3}, t_{2,\text{out},4}, t_{1,\text{in},4}$. The set $\mathcal{C}_1(n)$ contains an instance of class 2 and one instance of class 3 except when the object with class 3 is transported or kept in $c^2_{2,2}$; consistently, during that time, there is a detection of two instances of the class 3 inside $c^2_{2,2}$ (see $\mathcal{C}_2(n)$ for $t_{2,\text{in},3} < t \leq t_{2,\text{out},4}$). Since both classes of objects are made of the material 4 and the material 7, at every time it holds that $\mathcal{M}_1(n) = \mathcal{M}_2(n) = \{4, 7\}$. Both $\hat{m}_1(n)$ and $\hat{m}_2(n)$ have two jumps with magnitude $\overline{m} = 80$ grams, but the jumps occur at different time instants because the transported object of mass $\overline{m}$ exits and enters the nodes at different times. Specifically, $\hat{m}_k(n)$ increases when $\overline{m}$ enters the node $k$ and it decreases when $\overline{m}$ exits the node $k$. The jump discontinuities occur also in $\hat{F}^\psi_k$, but with a magnitude equal to the amount of material $\psi$ that exits or enters the node $k$. In this case, $\overline{m}$ is made of 50\% of plastic and 50\% of rubber, and hence, the magnitude of the jumps in $\hat{F}^\psi_k$ is equal to $\overline{m}/2 = 40$ grams for all the combinations of $\psi$ and $k$. Finally, the total detected mass $\hat{l}(n)$ has four jump discontinuities with magnitude equal to $\overline{m}$. The highest value of $\hat{l}(n)$ is 180 grams and it occurs before and after the transportation of the moving object; indeed, during transportation the set of units $\mathcal{U} = \{1,2\}$ does not detect the moving object because the object is inside a compartment with $k \not \in \mathcal{U}$. During transportation, the units detect a total of two objects, whereas they detect a total of three objects when there is no transportation. The transportation takes place for $t_{1,\text{out},3} < t \leq t_{2,\text{in},3}$ and for $t_{2,\text{out},4} < t \leq t_{1,\text{in},4}$.

}

\subsection{Prototype for Medical Materials}\label{sub:prototype}
\revision{This section covers the design of a prototype whose working principle is based on the framework in Section \ref{sec:theory}. Consider again the material network $\mathcal{N}$ in Fig. \ref{fig:consideredTMN}. In practice, we need to perform two tasks: 
\begin{enumerate}
\item{Sort different types of end-of-life products in the sorting facility, i.e., in $c^1_{1,1}$; specifically, in this case, the end-of-life products are inhalers.}
\item{Map and quantify the material masses inside the nodes in Fig. \ref{fig:consideredTMN} in real-time, i.e., estimate $m_1(n)$ and $m_2(n)$ in Equations (\ref{eq:mass1})-(\ref{eq:mass2}) and their geographical locations. This task can be seen as the answer to the following three questions: ``\emph{How much} material is inside the nodes of $\mathcal{N}$?'', ``\emph{What} materials are inside the nodes of $\mathcal{N}$?'', and ``\emph{Where} are the detected materials?''.}
\end{enumerate} 
}
It is important to note that Task 1 applies to the \emph{sorting compartment only}, i.e., $c^1_{1,1}$, whereas Task 2 operates at a \emph{systemic level}. Hence, hereinafter, we will refer to them as \emph{compartmental detection} and \emph{distributed detection}, respectively.

\subsubsection{Compartmental Detection}\label{sub:compartment} %
\begin{figure*}
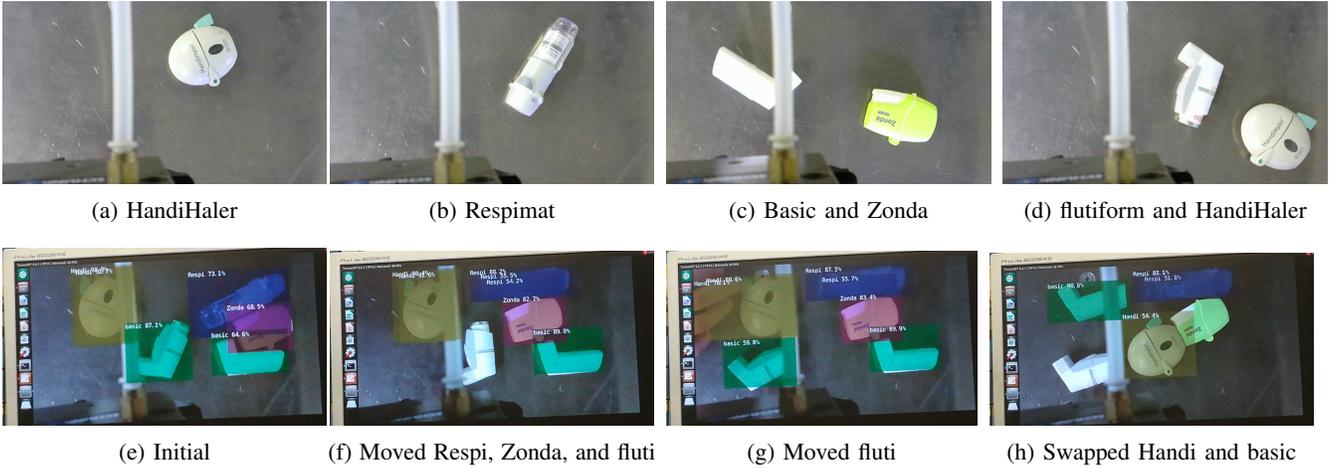

\begin{subfigure}{0.24\textwidth}
  \centering
  \includegraphics[width=.99\linewidth]{Figures/sample\_1}
  \caption{HandiHaler}
  \label{fig:Handi}
  \vspace{3mm}
\end{subfigure}%
\begin{subfigure}{0.24\textwidth}
  \centering
  \includegraphics[width=.99\linewidth]{Figures/sample\_2}
  \caption{Respimat}
  \label{fig:Respi}
  \vspace{3mm}
\end{subfigure}
\begin{subfigure}{0.24\textwidth}
  \centering
  \includegraphics[width=.99\linewidth]{Figures/sample\_3}
  \caption{Basic and Zonda}
  \label{fig:BandZ}
  \vspace{3mm}
\end{subfigure}
\begin{subfigure}{0.24\textwidth}
  \centering
  \includegraphics[width=.99\linewidth]{Figures/sample\_4}
  \caption{flutiform and HandiHaler}
  \label{fig:fAndH}
  \vspace{3mm}
\end{subfigure} \\ 
\begin{subfigure}{0.24\textwidth}
  \centering
  \includegraphics[width=.99\linewidth]{Figures/ODresult\_1}
  \caption{Initial}
  \label{fig:Init}
\end{subfigure}%
\begin{subfigure}{0.24\textwidth}
  \centering
  \includegraphics[width=.99\linewidth]{Figures/ODresult\_2}
  \caption{Moved Respi, Zonda, and fluti}
  \label{fig:MovedRZf}
\end{subfigure}
\begin{subfigure}{0.23\textwidth}
  \centering
  \includegraphics[width=.99\linewidth]{Figures/ODresult\_3}
  \caption{Moved fluti}
  \label{fig:Movedf}
\end{subfigure}
\begin{subfigure}{0.24\textwidth}
  \centering
  \includegraphics[width=.99\linewidth]{Figures/ODresult\_4}
  \caption{Swapped Handi and basic}
  \label{fig:SwappedHb}
\end{subfigure} 
\caption{Training samples ((a)--(d)) and tracking results ((e)--(h)).}
\label{fig:samplesAndRes}
\end{figure*}

Live object detection can enable the autonomous sorting of end-of-life inhalers in the node compartment $c^1_{1,1}$, which is a waste sorting facility. To design the detector, we collected and fully annotated 1000 images with the Pascal VOC format using the GUI of an NVIDIA Jetson Nano \cite{NvidiaODtutorial}. The images are distributed as follows: 160 with the basic inhaler (label: ``basic''), 160 with the HandiHaler model (label: ``Handi''), 160 with the Respimat model (label: ``Respi''), 160 with the Zonda model (label: ``Zonda''), 160 with the flutiform model (label: ``fluti''); this results in 800 images. We collected images of pairs of inhalers equally distributed so that each model appears with all the others the same number of times to avoid class imbalance. The possible combinations/pairs are $\frac{s!}{r!(s-r)!} = 10$, with $s = 5$ and $r = 2$. For each possible pair, we collected 20 images. Thus, there were 200 images of pairs of inhalers. These were added to the single-model images resulting in 1000 image samples in total. Sample images are shown in Figs. \ref{fig:Handi}--\ref{fig:fAndH}.

In addition, for each inhaler, we also annotated the pick point for picking it up with a vacuum gripper (label: ``pkPnt''). The pick point is essentially a small bounding box located in a position that is quite flat and internal to the device body so that the vacuum gripper is able to generate the vacuum between the inhaler surface and the suction cup. The dataset, which is named ``5IPP'' (5 Inhalers with Pick Points), has 6 classes: ``basic'', ``Handi'', ``Respi'', ``Zonda'', ``fluti'', and ``pkPnt''. \revision{A summary of the dataset is given in Table \ref{tab:5IPP}, where an entry on the main diagonal (e.g.,``basic-basic'') indicates the number of images depicting that inhaler model (e.g., model ``basic''), while an off-diagonal entry (e.g., ``basic-Zonda'') indicates the number of images depicting one inhaler per model (e.g., as in Fig. \ref{fig:BandZ}).}
\begin{table}
\centering
\begin{tabular}{c c c c c c |c} 
 & basic & Handi & Respi & Zonda & fluti & pkPnt \\ [0.5ex] 
 \hline
basic  & 160 & 20 & 20 & 20 & 20 & \multirow{5}{*}{\makecell{A bounding \\ box on each \\ inhaler}}\\ 
Handi & - & 160 & 20 & 20 & 20 & \\
Respi & - & - & 160 & 20 & 20 & \\
Zonda & - & - & - & 160 & 20 & \\  
fluti & - & - & - & - & 160  & \\ 
 \hline
\multicolumn{7}{c}{Summary: 6 classes, 1000 sample images}\\
\hline      
\end{tabular}
\caption{\revision{Number of sample images per each class of the 5IPP dataset.}}
\label{tab:5IPP}
\end{table}

On the same NVIDIA Jetson Nano, we trained a pre-trained SSD-MobileNet on 5IPP. The pre-training was performed on the MS COCO dataset \cite{lin2014microsoft}. We trained the model for 100 epochs with a batch size of 2 using the SGD optimizer with a learning rate of 0.01, momentum of 0.9, weight decay of $5 \times 10^{-4}$, and gamma of 0.1. We used 100\% of the samples for training \revision{to maximize the number of samples seen by the model, and hence, to maximize its generalization during deployment. A quantitative evaluation of the progress of a training with these settings is reported in Table \ref{tab:mAPvalues}, where it is visible that the values for the pkPnt class are significantly lower than for any other class. This is because the pkPnt class has a very small bounding box (i.e., $< 6 \text{ cm}^2$), which makes its recognition particularly difficult.}
\begin{table}
\centering
\begin{tabular}{c| c c c c c c |c} 
 Epoch & Basic & Handi & Respi & Zonda & fluti & pkPnt & mAP\\ [0.5ex] 
 \hline
 21 & 0.906 & 0.906  & 0.894 & 0.899 & 0.875 & 0.009 & 0.748 \\ 
 41 & 0.903 & 0.907 & 0.896 & 0.893 & 0.879 & 0.032 & 0.752 \\
 61 & 0.902 & 0.906 & 0.892 & 0.896 & 0.887 & 0.037 & 0.753 \\ 
 81 &  0.902 & 0.907 & 0.893 & 0.893 & 0.888 & 0.060 & 0.757 \\
 100 & 0.901 & 0.907 & 0.890 & 0.893 & 0.888 & 0.059 & 0.757 \\ [0.5ex] 
 \hline
\end{tabular}
\caption{\revision{Average precision per class of the 5IPP dataset and mean average precision (mAP) for SSD-MobileNet during a training of 100 epochs.}}
\label{tab:mAPvalues}
\end{table}

A qualitative performance evaluation is visible from the demo video\footnotemark[1] and from Figs. \ref{fig:Init}--\ref{fig:SwappedHb}, which are frames of the demo video in chronological order. Fig. \ref{fig:Init} shows the detection with a random initial configuration of all the inhalers. The percentage next to the class is the detection confidence. Note that the system works even if the training samples depicted either one or two objects only. The accuracy, however, is not perfect: there is a double detection of HandiHaler, the flutiform model is classified as basic, and the pick points are missing. A longer training ($>$ 100 epochs) could further increase the overall accuracy and mitigate these misdetections. Fig. \ref{fig:MovedRZf} shows the detection after the Respimat, flutiform, and Zonda models have been moved. The system still works although the detection of flutiform is lost, but it is recovered after it has been moved again (Fig. \ref{fig:Movedf}). Finally, the HandiHaler and the basic models were swapped in Fig. \ref{fig:SwappedHb} and the detector correctly tracked the change.                

\subsubsection{Distributed Detection}\label{sub:distributed} %
To automatically map and quantify the constituent materials of the inhalers, we simulated placing an MMU in each node compartment, i.e., in $c^1_{1,1}$ and $c^2_{2,2}$ (Fig. \ref{fig:consideredTMN}). Each MMU consists of an NVIDIA Jetson Nano microprocessor running the detector covered in Section \ref{sub:compartment} and connected to a Logitech C270 webcam. The two GPS modules, tested but not yet integrated into the whole system, are the model NEO-6M-0-001 with SMA connector, which can be connected to a Nano serial port via a CP2102-USB-to-TTL-UART serial converter module. 

\begin{figure}
\includegraphics[width=0.48\textwidth]{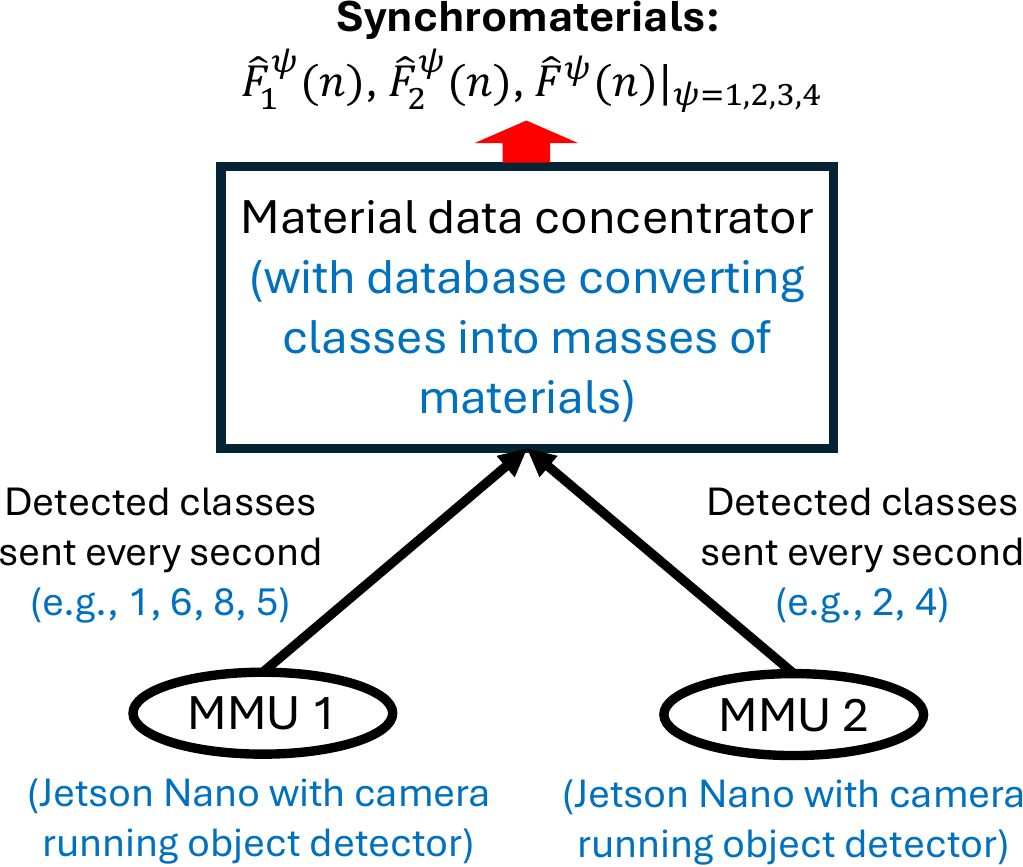}
\centering
\caption{\revision{Communication diagram of the two-unit system.}}
\label{fig:commDiag}
\end{figure}

The frames captured by each webcam are processed by one neural detector with a speed of approximately 12-22 frames per second. Every second, each unit sends the detected classes to the \emph{material data concentrator} (Fig. \ref{fig:commDiag}), which is a laptop in this prototype (note the analogy with the \emph{phasor data concentrator} \cite{dwivedi2023dynamopmu}).  
The communication between the units and the data concentrator is via Ethernet cables. Specifically, the laptop and the two Jetson Nano microprocessors are connected through a network switch, model Netgear FS105 with 5 ports. The timing of each unit is given by the Nano internal clock. 

\begin{figure}
\begin{subfigure}{0.48\textwidth}
\includegraphics[width=\textwidth]{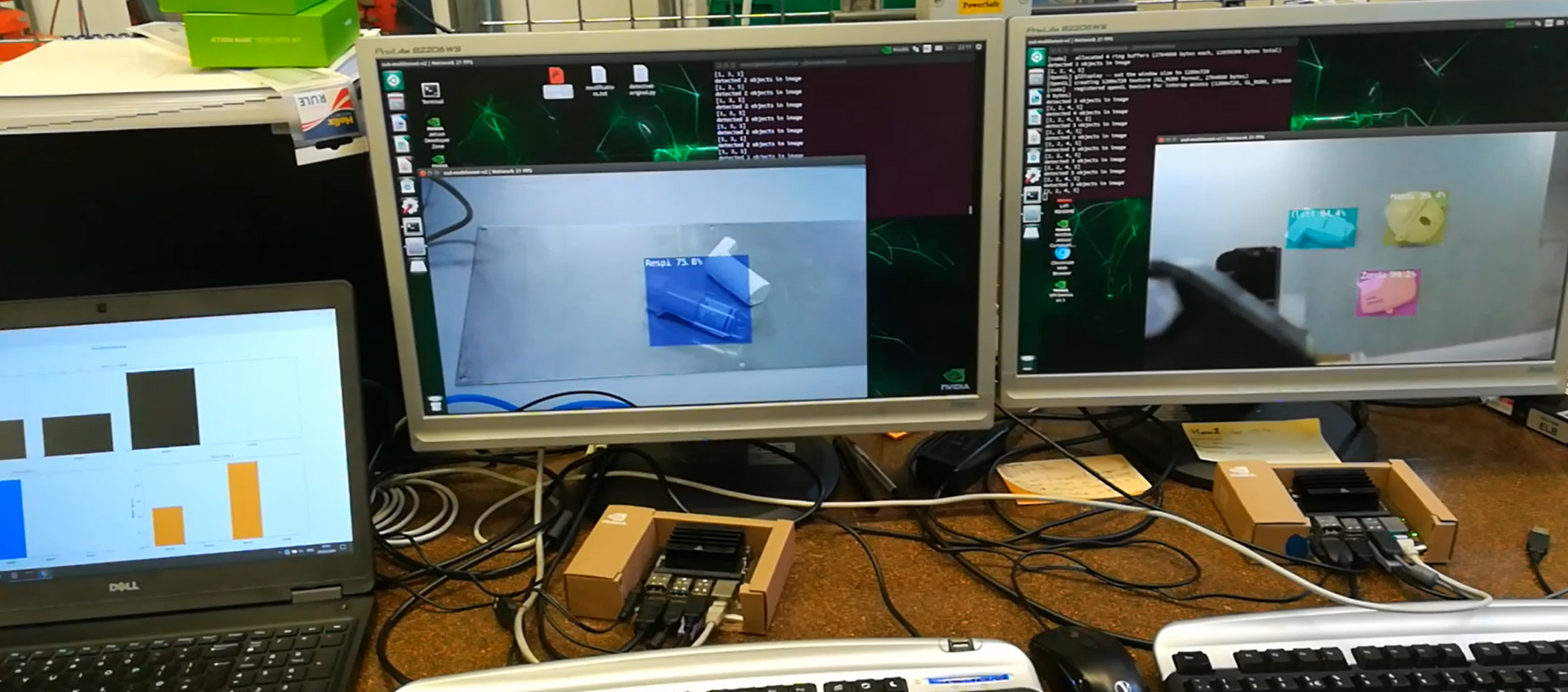}
\centering
\caption{The two MMUs, their detections, and the synchromaterials computed by the material data concentrator (laptop).}
\label{fig:wholeSystem}
\vspace{3mm}
\end{subfigure} \\
\begin{subfigure}{0.48\textwidth}
\includegraphics[width=\textwidth]{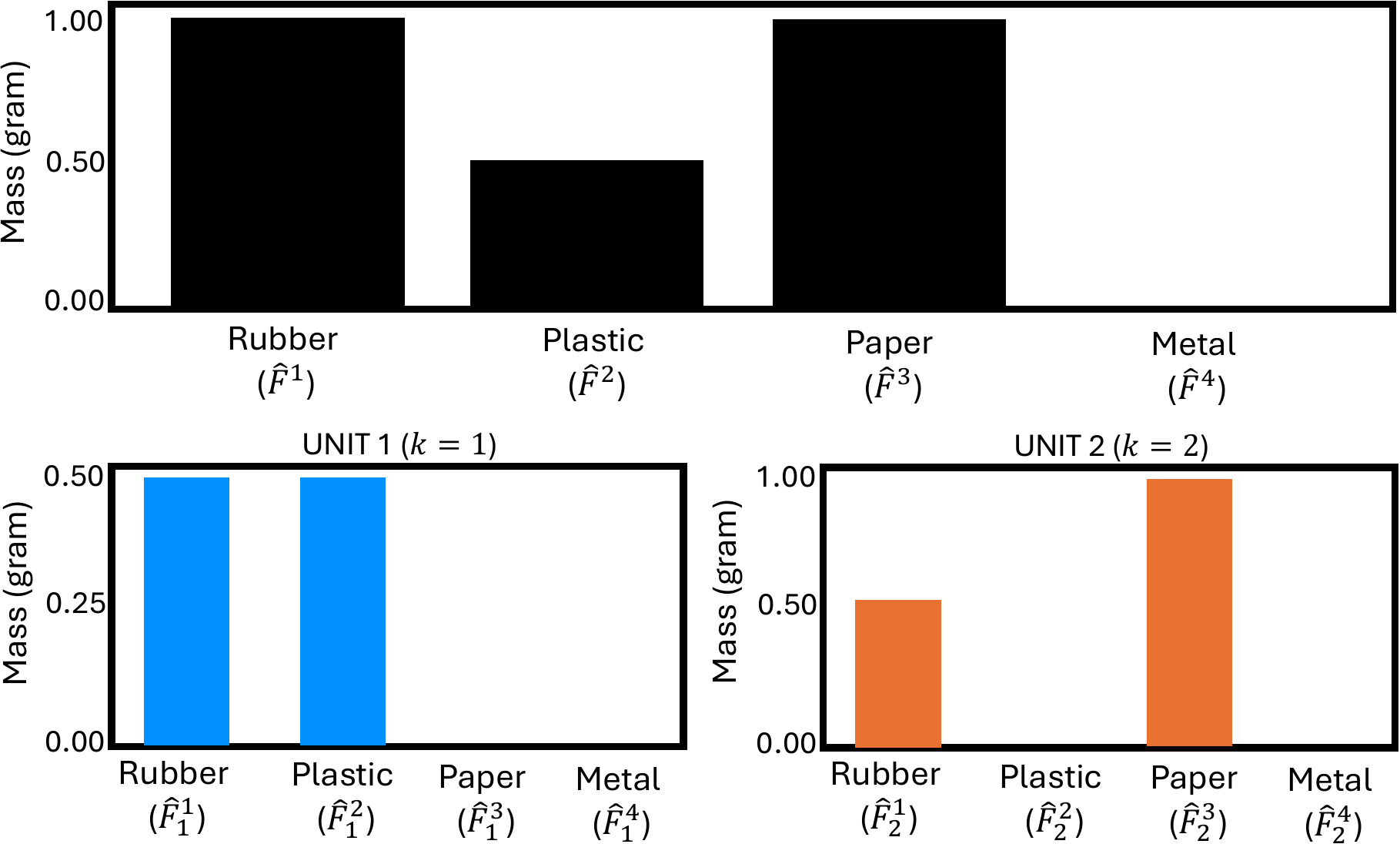}
\centering
\caption{\revision{Detail of the synchromaterials displayed by the material data concentrator and updated every second.}}
\label{fig:barCharts}
\end{subfigure} 
\caption{Live measures of synchromaterials.}
\label{fig:synchromat}
\end{figure}

The synchromaterials are displayed in bar charts to facilitate the reading to the material resources engineer or manager as in Fig. \ref{fig:synchromat}. A demo video is available\footnotemark[1]. The current system works with only one instance per object class, e.g., by considering only one Zonda inhaler even when two are detected. The current list of materials is merely to show the working principle of the system as visible in Fig. \ref{fig:barCharts}, where the $x$-axis labels are ``rubber'', ``plastic'', ``paper'', and ``metal'': there is no paper in these inhalers. Note also in Fig. \ref{fig:barCharts} that the chart at the top (with bars in black) is the sum of the masses from Unit 1 and Unit 2 (bars at the bottom in blue and orange, respectively). The GPS module can provide the location of the unit it is connected to, and hence, the location of the detected materials.  

\section{Conclusion}\label{sec:Conclusions}
This paper proposed real-time synchronized object detection to simultaneously automate sorting, mapping, and quantification of materials. The constituent materials of five models of inhalers were considered as the case study. The working principle of the system is detailed to demonstrate its applicability and scalability. 

This work is a first step towards wide-area autonomous materials monitoring. A large-scale computational infrastructure based on MMUs can facilitate the implementation of a circular economy by improving decision making about material resources management. The role that MMUs can play in material networks is analogous to the role that PMUs have in electrical networks today. 

For sorting, future work could be to improve the accuracy of the detector \revision{in order to detect also the pick points, and explore alternatives such as YOLO \cite{wang2024gold} and vision transformers \cite{zhao2024detrs}. In addition, the NVIDIA Jetson Xavier or the Google Coral Dev board could be used and compared with the Jetson Nano kit.} For autonomous mapping and quantification, future work could be to synchronize the MMUs using a common clock such as the pulse-per-second (PPS) signal provided by the GPS, enable the system to work with multiple instances per class, compile a more precise list of materials for each object class, and address the MMU placement problem (analogous to the PMU placement problem).

\section*{Acknowledgment}
The authors gratefully thank the ReMed Team (\url{https://www.remed.uk/team}) for \revision{their insightful discussions}.

\Urlmuskip=0mu plus 1mu\relax
\bibliographystyle{IEEEtran}
\bibliography{references}

\newpage

\vfill

\end{document}